\pdfoutput=1

\documentclass[11pt]{article}

\usepackage[]{emnlp2021}

\usepackage{times}
\usepackage{latexsym}

\usepackage[T1]{fontenc}

\usepackage[utf8]{inputenc}

\usepackage{microtype}
\usepackage{graphicx} 
\usepackage{multirow}
\usepackage{comment}
\usepackage{tabularx}
\usepackage{booktabs}
\usepackage{latexsym}
\usepackage{amsmath}
\usepackage{amssymb}
\usepackage{mathrsfs}
\usepackage{enumerate}
\usepackage{bm}
\usepackage{paralist}
\usepackage{subfigure}
\usepackage{caption}
\usepackage{fancyhdr}
\usepackage{courier}
\usepackage{color,xcolor,colortbl}
\usepackage{array}
\usepackage{color}
\usepackage{enumitem}
\usepackage[ruled,vlined]{algorithm2e}
\definecolor{c1}{rgb}{0.12, 0.46, 0.70}
\definecolor{c2}{rgb}{1.0, 0.50, 0.05}
\definecolor{c3}{rgb}{0.17,0.63,0.17}

\newcommand{\textcite}[1]{\citeauthor{#1} (\citeyear{#1})}

\title{Graphine: A Dataset for Graph-aware Terminology Definition Generation}

\author{Zequn Liu$^1$, Shukai Wang$^1$, Yiyang Gu$^1$, Ruiyi Zhang$^1$, Ming Zhang$^1$\thanks{$^*$Corresponding author}, Sheng Wang$^{2*}$\\
  $^1$Department of Computer Science, School of EECS, Peking University, Beijing, China \\
  $^2$Paul G. Allen School of Computer Science and Engineering, University of Washington, Seattle, WA\\
   \texttt{{zequnliu,shukaiwang,yiyanggu,zhangruiyi,mzhang\_cs}@pku.edu.cn}\\
  \texttt{swang@cs.washington.edu}
  }
\begin{document}
\maketitle
\begin{abstract}
Precisely defining the terminology is the first step in scientific communication. Developing neural text generation models for definition generation can circumvent the labor-intensity curation, further accelerating scientific discovery. Unfortunately, the lack of large-scale terminology definition dataset hinders the process toward definition generation. In this paper, we present a large-scale terminology definition dataset Graphine covering 2,010,648 terminology definition pairs, spanning 227 biomedical subdisciplines. Terminologies in each subdiscipline further form a directed acyclic graph, opening up new avenues for developing graph-aware text generation models. We then proposed a novel graph-aware definition generation model Graphex that integrates transformer with graph neural network. Our model outperforms existing text generation models by exploiting the graph structure of terminologies. We further demonstrated how Graphine can be used to evaluate pretrained language models, compare graph representation learning methods and predict sentence granularity. We envision Graphine to be a unique resource for definition generation and many other NLP tasks in biomedicine.\footnote{Our Dataset is available at \url{https://zenodo.org/record/5320310\#.YSxHgI77Q2w}. Our code is available at \url{https://github.com/zequnl/Graphex}}
\end{abstract}
\section{Introduction}
Obtaining the definition is the first step toward understanding a new terminology. The lack of precise terminology definition poses great challenges in scientific communication and collaboration \cite{oke2006towards,cimino1994knowledge}, which further hinders new discovery. This problem becomes even more severe in emerging research topics \cite{baig2020chronic,baines2020defining}, such as COVID-19, where curated definitions could be imprecise and do not scale to rapidly proposed terminologies. 

Neural text generation \cite{bowman2015generating,transformer,sutskever2014sequence,song2020learning} could be a plausible solution to this problem by generating definition text based on the terminology text. Encouraging results by neural text generation have been observed on related tasks, such as paraphrase generation \cite{li2020unsupervised}, description generation \cite{cheng2020ent}, synonym generation \cite{gupta2015unsupervised} and data augmentation \cite{malandrakis2019controlled}. However, it remains unclear how to generate definition, which 
comprises concise text in the input space (i.e., terminology) and longer text in the output space (i.e., definition). Moreover, the absence of large-scale terminology definition datasets impedes the progress towards developing definition generation models.
\begin{figure}[!t]
\includegraphics[width=0.45\textwidth]{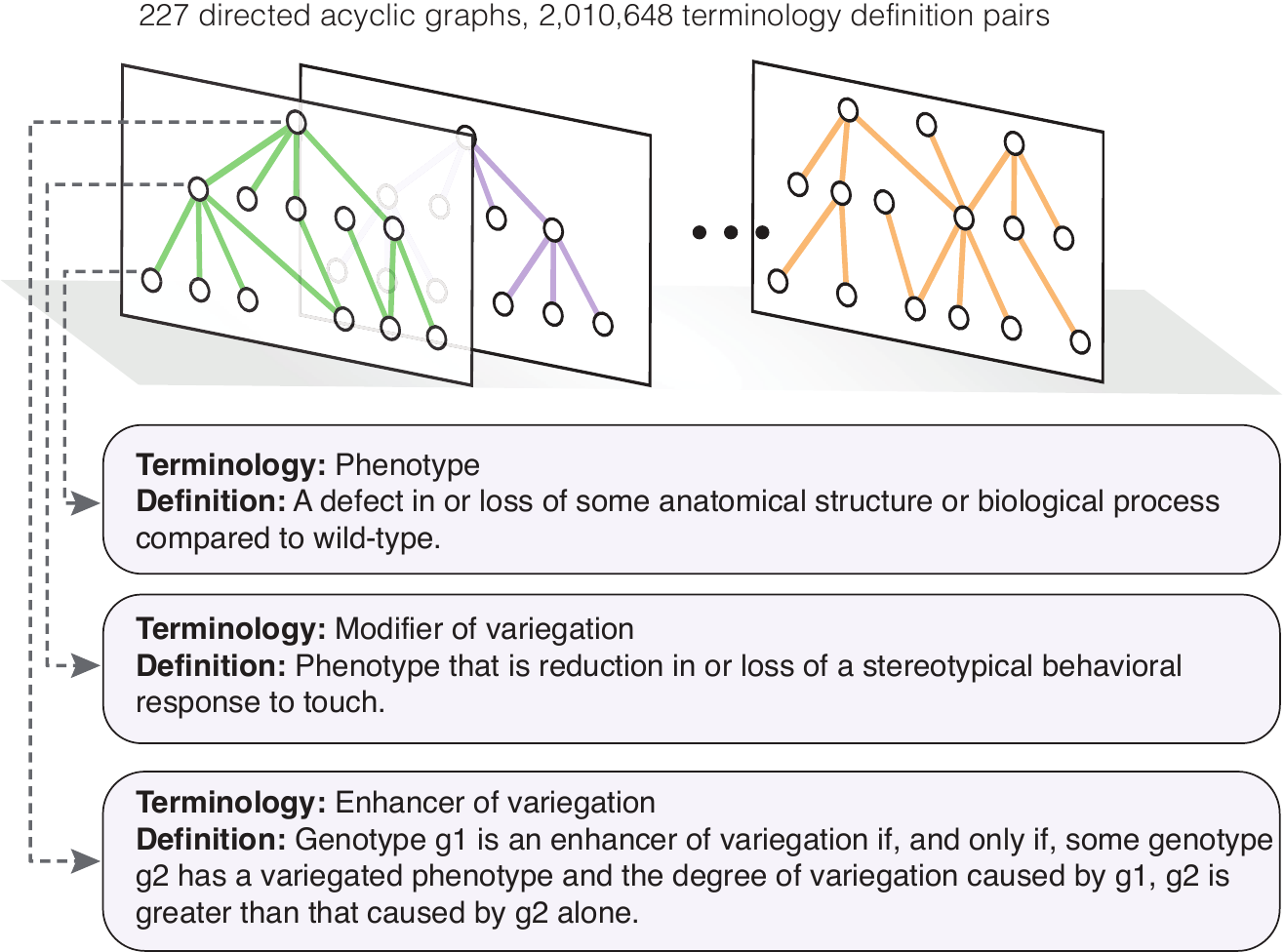}
\caption{Graphine dataset contains 2,010,648 terminology definition pairs organized in 227 directed acyclic graphs. Each node in the graph is associated with a terminology and its definition. Terminologies are organized from coarse-grained ones to fine-grained ones in each graph.}
\label{fig:1}
\end{figure}

Despite these challenges, scientific terminologies often form a directed acyclic graph (DAG), which could be helpful in definition generation. Each DAG organizes related terminologies from general ones to specific ones with different granularity levels (\textbf{Figure \ref{fig:1}}). These DAGs have proved to be useful in assisting disease, cell type and function classification \cite{wang2020unifying,song2020generalized,wang2015exploiting} by exploiting the principle that nearby terms on the graph are semantically similar \cite{altshuler2000guilt}. Likewise, terminologies that are closer on this DAG should acquire similar definitions. Moreover, placing a new terminology in an existing DAG requires considerably less expert efforts than curating the definition, further motivating us to generate the definition using the DAG.

In this paper, we collectively advance definition generation in the biomedical domain through introducing a terminology definition dataset Graphine and a novel graph-aware text generation model Graphex. Graphine encompasses 2,010,648 terminology definition pairs encapsulated in 227 DAGs. These DAGs are collected from three major biomedical ontology databases \cite{obo,bioportal,ols}. All definitions are curated by domain experts. Our graph-aware text generation model Graphex utilizes the graph structure to assist definition generation based on the observation that nearby terminologies exhibit semantically similar definitions.

Our human and automatic evaluations demonstrate the substantial improvement of our method on definition generation in comparison to existing text generation methods that do not consider the graph structure. In addition to definition generation, we illustrate how Graphine opens up new avenues for investigating other tasks, including domain-specific language model pretraining, graph representation learning and a novel task of sentence granularity prediction. Finally, we present case studies of a failed generation by our method, pinpointing directions for future improvement. To the best of our knowledge, Graphine and Graphex build up the first large-scale benchmark for terminology definition generation, and can be broadly applied to a variety of tasks.

\begin{figure}[!t]
\includegraphics[width=0.5\textwidth]{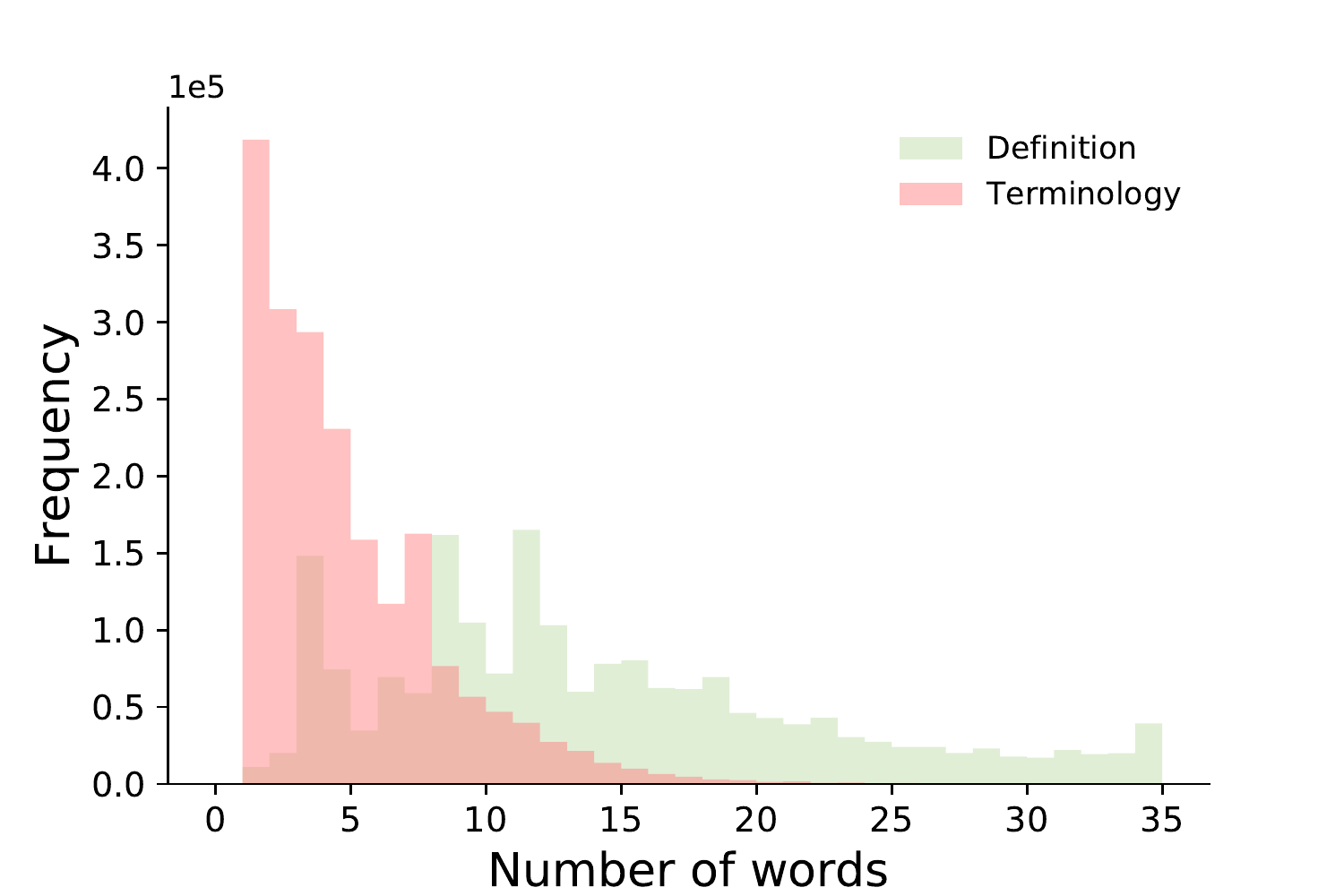}
\caption{Bar plot showing the comparison between the number of words in the definition and in the terminology in Graphine.}
\label{fig3}
\end{figure}


\section{Graphine Dataset}
\subsection{Data collection and statistics}
We collect 2,010,648 biomedical terminology definition pairs from three major biomedical ontology databases, including Open Biological and Biomedical Ontology Foundry (OBO)~\cite{obo}, BioPortal~\cite{bioportal} and EMBL-EBI Ontology Lookup Service (OLS)~\cite{ols}, spanning diverse biomedical subdisciplines such as cellular biology, molecular biology and drug development. For the definition that span multiple sentences, we only consider the first sentence.

Even though these large-scale terminology definition pairs have already presented a novel resource for definition generation, one unique feature of our dataset is the graphs among terminologies. In particular, we construct a DAG for each biomedical subdiscipline using `is a' relationship from the original data. As a result, each terminology belongs to one DAG, where the node is associated with a terminology and its definition and the edge links from a general terminology to a specific one. We reduce the number of DAGs from 499 to 227 by merging DAGs that appear in more than one database. 

We notice substantial amount of missing definitions in the original collection, confirming the importance of computationally generating definition. In 81 out of 499 DAGs, more than 50\% of terminologies does not have any definition. We thus exclude terminologies that do not have a curated definition. We further observed a substantial discrepancy between the number of words in the terminology and the number of words in the definition. The average number of words in the terminology is 4.55, which is much lower than the 15.58 average number of words in the definition (\textbf{Figure \ref{fig3}}). This discrepancy could pose great challenges to text generation model. We seek to alleviate it using graph neighbor's terminology and definition.

\begin{figure*}[!t]
\centering
\includegraphics[width=1\textwidth]{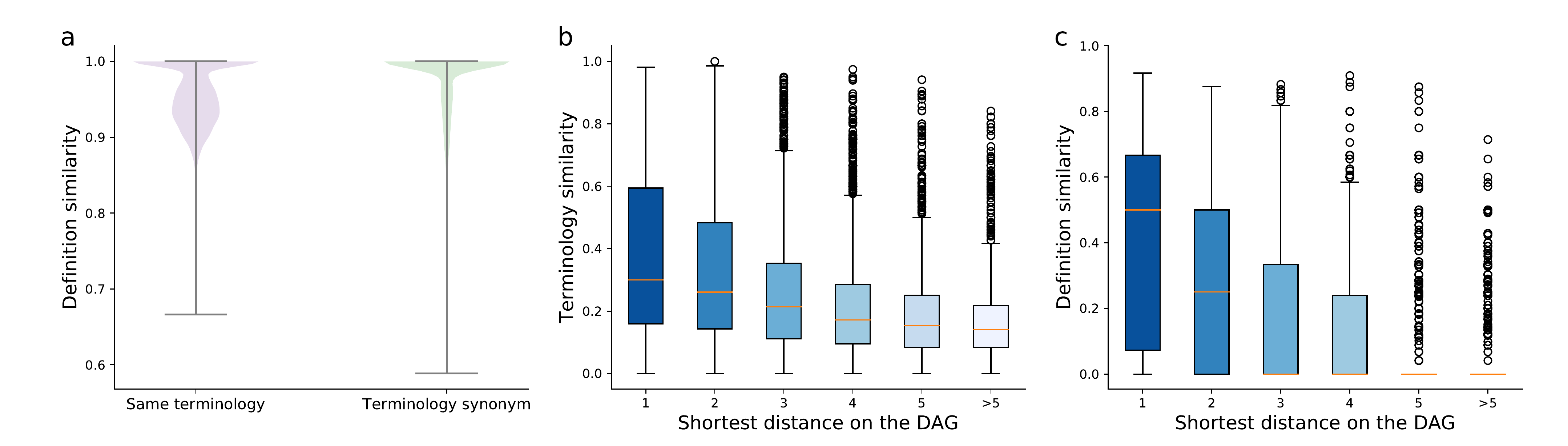}
\caption{Analysis of Graphine. a, Violin plot showing the definition similarity between the same terminology and the terminology synonym curated by different experts. b,c, Box plots showing the terminology similarity (b) and the definition similarity (c) between nodes of different shortest distances on the DAG.}
\label{fig:5}
\end{figure*}

\subsection{Data analysis}\label{sec:analysis}
All definitions in our datasets are curated by domain experts, assuring the high-quality. Reassuringly, we investigate the consistency between expert curation by comparing the definitions of the same terminology from different DAGs (e.g., \textit{material maintenance} appears in both \textit{obi} and \textit{chmo}). Different DAGs are curated by different domain experts in our dataset. We observed a remarkable cosine similarity of 0.96 between definitions of the same terminology (\textbf{Figure \ref{fig:5}a}). We next examine the definitions of 67,257 terminology synonym pairs that presents in different DAGs. Synonyms are also curated by domain experts in the original databases. We again observed prominent cosine similarity 0.97, assuring the consistency between expert curation.

To examine the quality of the graph structure, we study the consistency between graph-based terminology similarity and text-based terminology similarity. Graph-based terminology similarity is calculated using the shortest distance on the graph. Text-based similarity is calculated using BLEU score \cite{bleu} between two terminologies. We observed strong agreement between these two similarity scores (\textbf{Figure \ref{fig:5}b}). This agreement is even more substantial between graph-based terminology similarity and text-based definition similarity (\textbf{Figure \ref{fig:5}c}). Collectively, these results indicate that nearby nodes exhibit similar terminologies and definitions, suggesting the opportunity to improve definition generation using the graph structure. 

\section{Graph-aware Definition Generation: Task and Model}
\subsection{Problem Definition}
Our goal is to generate the definition text according to the terminology text. Meanwhile, terminologies form a DAG, which could be used to assist definition generation. More precisely, the input is a directed acyclic graph $G = \left({V},{E},{T},{D}\right)$, where ${V}=\{v_i\}$ is the set of nodes and ${E} \subseteq {V} \times {V}$ is the set of edges. Each node $v_i$ is associated with a terminology $t_i \in {T}$ and a definition $d_i \in {D}$. $t_i$ and $d_i$ are both token sequences defined as $t_i \triangleq \left<t_i^1, t_i^2,\ldots, t_i^{n_{t_i}}\right>$ and $d_i \triangleq \left<d_i^1, d_i^2,\ldots, d_i^{n_{d_i}}\right>$, where $t_i^j \in {C}$, $d_i^j \in {C}$ and ${C}$ is the vocabulary. In practice, the terminology is often a phrase and the definition is a sentence. Therefore, $n_{d_i}$ is much larger than $n_{t_i}$. 

We consider a transductive learning setting where $V$ composes of ${V}_{train}$ and ${V}_{test}$. ${V}_{train}$ is the set of nodes that have both terminologies and definitions. ${V}_{test}$ is the set of nodes that only have terminologies. The goal of graph-aware definition generation is to generate $d_i$ for $v_i \in {V}_{test}$ according to both the terminology $t_i$ and the graph $G$. Although each graph $G$ in Graphine is a DAG, our method can be applied to any kind of graphs. 

The proposed definition generation task is distinct from conditional text generation and machine translation due to the presence of this graph $G$. $G$ makes it possible to transfer knowledge between terminologies based on our previous observation that nearby nodes on the graph have similar definitions. We thus aim at propagating terminology and definition using the graph structure to enhance definition generation.
 
\subsection{Model}
\begin{figure}[!t]
\includegraphics[width=0.5\textwidth]{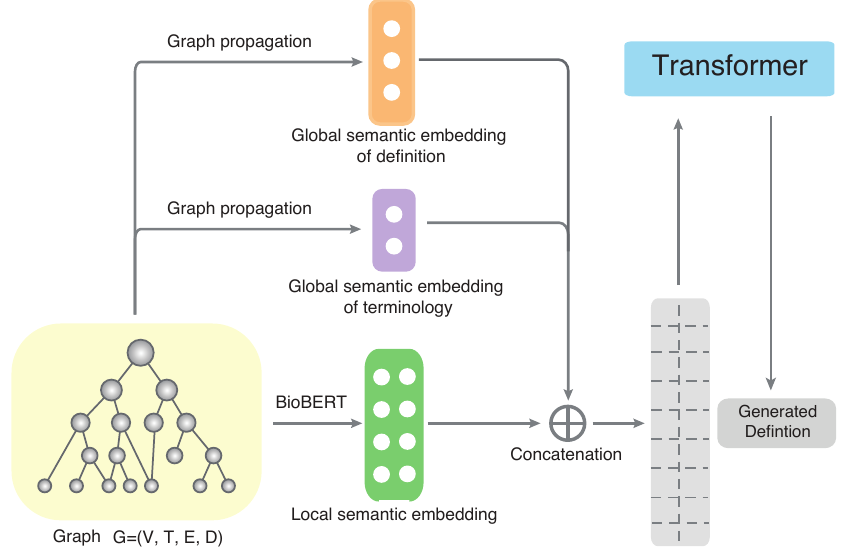}
\caption{Flowchart of Graphex. Graphex considers the graph structure during definition generation by concatenating the global semantic embeddings and the local semantic embedding.}
\label{fig:model}
\end{figure}

We propose a graph-aware definition generation approach Graphex that generates definition based on the global semantic embedding and the local semantic embedding using a two-stage approach (\textbf{Fig. \ref{fig:model}}). At the first stage, global semantic embeddings are calculated through propagating terminology and definition on the graph. At the second stage, the local semantic embedding is obtained by embedding the specific terminology. Finally, Graphex generates the definition $d_i$ by using the concatenation of global and local semantic embeddings as the input to a Transformer \cite{transformer}. 

\subsubsection{Encoding global semantic via graph propagation} \label{graph embedding}
At the first stage, we obtain two global semantic embedding $\boldsymbol g_i^t$ and $\boldsymbol g_i^d$ of each node $v_i$ through propagating terminology and definition on the graph, respectively. In particular, we follow a previous work \cite{kotitsas2019embedding} to calculate $\boldsymbol g_i^t$ and $\boldsymbol g_i^d$ using a bidirectional GRU-based neural network, which aggregates the embeddings of individual words in $t_i$ as the node features of the node $v_i$ and then smooths node features based on random walk. 

To encode the network structure, we sample $m$ random walk paths of fixed length $k$ starting from each node \cite{grover2016node2vec}. The $r$-th random walk starting from the node $v_i$ is denoted as $P_{v_i,r} = \left<p_{1,r}, p_{2,r},...,p_{k,r}\right>$ ($r=1, \dots, m$), where $p_{1,r}=v_i$. We then learn two embeddings $\boldsymbol w_i$ and $\boldsymbol u_i$ for each node $v_i$ based on the arriving probability calculated from these sampled random walk paths. In particular, the predicted probability of arriving the node $v_j$ through the walk $P_{v_i,r}$ is defined as:
\begin{equation}
p(v_j|v_i)=
\dfrac{\exp(\boldsymbol u_j^T \boldsymbol w_i)}{\sum_{\boldsymbol v_k \in V} \exp(\boldsymbol u_k^T \boldsymbol w_i)}.
\label{eq:softmax}
\end{equation}
Here, $\boldsymbol w_i$ is the feature embedding and $\boldsymbol u_i$ is the context embedding for node $v_i$. 

Instead of training $\boldsymbol w_i$ and $\boldsymbol u_i$ solely based on the network structure, we use text feature from $t_i$ to regularize them. We define $\boldsymbol q(c)$ and $\boldsymbol h(c)$ to be the two separate trainable word embeddings for each token $c$ in the vocabulary $C$.
Then $\boldsymbol q(t_i^k)$ and $\boldsymbol h(t_i^k)$ are the trainable word embeddings of the $k$-th token in the terminology $t_i$. We use a shared bidirectional GRU network to encode $t_i$ into $\boldsymbol u_i$ and $\boldsymbol w_i$ as:
\begin{eqnarray}
 {\boldsymbol u_{i}^f} & = & {GRU_{f}}(\boldsymbol q(t_i^1),\dots,\boldsymbol q(t_i^n)) \\
 {\boldsymbol u_{i}^b} & = & {GRU_{b}}(\boldsymbol q(t_i^1),\dots,\boldsymbol q(t_i^n)) \\
 \boldsymbol u_i & = & Max\_pooling({\boldsymbol u_{i}^b} + {\boldsymbol u_{i}^f}) \\
 {\boldsymbol w_{i}^f} & = & {GRU_{f}}(\boldsymbol h(t_i^1),\dots,\boldsymbol h(t_i^n)) \\
 {\boldsymbol w_{i}^b} & = & {GRU_{b}}(\boldsymbol h(t_i^1),\dots,\boldsymbol h(t_i^n)) \\
 \boldsymbol w_i & = & Max\_pooling({\boldsymbol w_{i}^f} + {\boldsymbol w_{i}^b}).
\end{eqnarray}

The loss function at the first stage is defined as:
\begin{equation}
 L_1 = - \sum_{v_i \in {V}} \sum_{r=1}^{m} \sum_{j=2}^k 
  \log p(v_{j,r}|v_{1,r} = v_i)
\label{eq:loss}
\end{equation}
After minimizing this loss function, $\boldsymbol g_i^t$ is obtained by concatenating $\boldsymbol w_i$ and $\boldsymbol u_i$, which represents the global semantic of node $v_i$ using the terminology. Likewise, we can obtain $\boldsymbol g_i^d$ by first encoding $d_i$ into the feature embedding $\boldsymbol w'_i$ and the context embedding $\boldsymbol u'_i$, and then concatenating them. For node that does not have the definition (i.e., $v_i \in {V}_{test}$), we generate a $\boldsymbol d'_i$ as replacement by using $t_i$ as input to a Transformer trained on other terminology definition pairs.

\subsubsection{Fusing local and global semantic for definition generation}
At the second stage, we generate the definition $d_i$ for node $v_i$ conditioned on both the local semantic $\boldsymbol l_i$ and the global semantic $\boldsymbol g_i^t$ and $\boldsymbol g_i^d$. The local semantic $\boldsymbol l_i$ is obtained by embedding $t_i$ using BioBERT~\cite{lee2020biobert}. We also examined other BERT-based models in the experiments. Let $P(d_i|\boldsymbol l_i, \boldsymbol g_i^t, \boldsymbol g_i^d;\boldsymbol \theta)$ be the transformer model parameterized by $\boldsymbol \theta$. The loss function at the second stage is defined as
\begin{equation}
 L_2 = - \sum_{i \in |V|}
  \log P(d_i|\boldsymbol l_i,\boldsymbol g_i^t, \boldsymbol g_i^d;\boldsymbol \theta).
 \label{eq:bertloss}
\end{equation}

\section{Experimental Results}
\subsection{Experimental setup}
We conduct experiments using DAGs included in the OBO database. To study the effect of graph structures, we only consider graphs that show a high correlation between the graph-based similarity and the text-based definition similarity as measured in Section \ref{sec:analysis}. Only definitions of the training data are used to calculate the correlation. We split the terminology definition pairs into 70\% training, 10\% validation and 20\% test. The data split and model training are done within each DAG separately.

\begin{table*}[]
\small
\centering
\resizebox{\linewidth}{!}{
\begin{tabular}{l|p{3mm}p{3mm}p{9mm}p{9mm}p{9mm}p{9mm}p{12mm}p{7mm}p{10mm}}
\hline
\textbf{Model}&\textbf{TG}&\textbf{DG}&\textbf{BLEU1}&\textbf{BLEU2}&\textbf{BLEU3}&\textbf{BLEU4}&\textbf{METEOR}&\textbf{NIST}&\textbf{Human}\\
\hline
Seq2Seq &&& 21.52	&14.47	&10.82	&8.56	&9.42		&0.69&0.87\\
CVAE&&&20.05&13.48&10.02&8.23&8.97&0.67&0.83\\
Transformer &&&31.91	&25.09	&21.26	&18.70	&15.81		&1.06&1.01\\
\hline
Our Model w/o TG&&\checkmark&33.81&26.23&22.16&19.26&16.55&1.10&1.09\\
Our Model w/o DG &\checkmark&&32.47	&25.52	&21.73	&19.30	&16.17		&1.11&1.06\\
Our Model &\checkmark&\checkmark& \textbf{34.35}	&\textbf{26.97}	&\textbf{22.99}	&\textbf{20.21}	&\textbf{16.57}		&\textbf{1.15}&\textbf{1.12}\\
\hline
\end{tabular}}
 \caption{Comparison on the performance of definition generation using automatic and human evaluation. TG (DG) refers to propagation on terminologies (definitions).}
 \label{tab:result}
\end{table*}
\begin{table*}[]
\small
\centering
\begin{tabular}{l|p{130mm}}
\hline
\textbf{Terminology:}&estuarine tidal riverine open water pycnocline\\
\textbf{True definition:}&an \textcolor{blue}{estuarine open water pycnocline} which is composed primarily of fresh tidal water\\
\textbf{Parent definition:}&a pycnocline which is part of an \textcolor{blue}{estuarine} water body, spanning from a fiat boundary where \textcolor{blue}{the estuary bed below the water column} reaches a depth of 4 meters until the end of the estuary most distal from the coast\\
\textbf{Graphex:}&an \textcolor{blue}{estuarine water} which extends from an estuarine pycnocline or mid - depth to the \textcolor{blue}{estuary bed} and from a fiat boundary where \textcolor{blue}{the estuary bed below the water column}\\
\textbf{Transformer:}&an area of a planet's surface which is primarily covered by UNK herbaceous vegetation and where the underlying soil or\\
\hline
\textbf{Terminology:}&increased eye tumor incidence\\
\textbf{True definition:}&greater than the expected number of tumors originating in the eye \textcolor{blue}{in a given population in a given time period}\\
\textbf{Child definition:}&greater than the expected number of neoplasms in the retina, usually in the form of a distinct mass, \textcolor{blue}{in a specific population in a given time period}\\
\textbf{Graphex:}&greater than the expected number of neoplasms in the gastric tissue usually in the form of a distinct mass , \textcolor{blue}{in a specific population in a given time period}\\
\textbf{Transformer:}&greater than the expected number of UNK in the lung , usually in the form of a distinct mass\\
\hline
\end{tabular}
 \caption{Comparison between definitions generated by Graphex and the best baseline Transformer. True definitions of the nearby node are also listed to illustrate the effect of considering graph structures.}
 \label{tab:case}
\end{table*}

We compare our method with three conventional conditional text generation models: \textbf{Seq2Seq} \cite{seq2seq-attention}, \textbf{CVAE} \cite{yan2016attribute2image} and \textbf{Transformer} \cite{transformer}. All of them take the terminology as the input and the definition as the output. Since none of them considers the graph structure, our comparison could reveal the importance of considering graph structures. We further implement two variants of our model to investigate the impact of propagating definition on the graph and propagating terminology on the graph. In particular, \textbf{Our Model w/o TG} is the Graphex framework that does not incorporate the terminology-derived global semantic embedding $\boldsymbol g_i^t$ in eq. \ref{eq:bertloss}. \textbf{Our Model w/o DG} is the Graphex framework that does not incorporate the definition-derived global semantic embedding $\boldsymbol g_i^d$ in eq. \ref{eq:bertloss}.

We used the same pretrained language model for all the competing methods. We chose BioBERT as it achieved the best performance among different pretrained language models. LSTM is used as the encoder and the decoder of \textbf{Seq2seq} and \textbf{CVAE} and the dimensions of the word embedding and the hidden state are set to 768. The dimensions of the word embedding and the hidden state of \textbf{Transformer} are also set to 768. In our model, we used the default hyperparameters in \cite{kotitsas2019embedding} in the first stage and use the same structure as \textbf{Transformer} baseline in the second stage. The dimensions of $g_i^d$ and $g_i^t$ are 768. All the models were trained using the same data splits. 

We used Graphex as a benchmark to compare pretrained language models on Graphine. We use \textbf{BERT}~\cite{bert}, \textbf{RoBERTa}~\cite{roberta}, \textbf{SciBERT}~\cite{scibert}, \textbf{PubMedBERT}~\cite{pubmedbert} and \textbf{BioBERT}~\cite{lee2020biobert} to provide the pretrained word embeddings for Graphine respectively. \textbf{SciBERT} fine-tunes BERT on scientific data. \textbf{PubMedBERT} and \textbf{BioBERT} are domain-specific BERTs on biomedical domain. The word embedding dimensions are all set to 768.

We compare different graph embedding methods, \textbf{GCN} \cite{gcn}, \textbf{HGCN} \cite{hgcn} and \textbf{GAT} \cite{gat} on our dataset. The AUC and AP of link prediction are used to evaluate the quality of graph embedding. We compare the three graph neural network methods with Euclidean embeddings and Poincare embeddings methods, \textbf{Euclidean} and \textbf{PoincareBall} \cite{nickel2017poincar}, and feature-based methods, \textbf{HNN} \cite{hnn} and \textbf{MLP}. We follow the default hyperparameter settings in \cite{hgcn}

We perform both automatic evaluation and human evaluation. For automatic evaluation, we used six standard metrics including \textbf{BLEU1-4}~\cite{bleu}, \textbf{METEOR}~\cite{banerjee2005meteor} and \textbf{NIST}~\cite{doddington2002automatic}. \textbf{BLEU1-4} measures the n-gram overlap between the generated sentence and the target sentence. \textbf{METEOR} improves BLEU by considering synonyms when comparing unigrams and using F1 instead of precision. \textbf{NIST} reweights words by frequency when matching n-grams to adjust the contribution of common words like "is". For human evaluation, we recruited 3 annotators to score the generated sentences of each method for 50 terminologies. Annotators are requested to grade each generated definition as 0 (bad), 1 (fair) and 2 (good). 

\subsection{Graphex improves definition generation by considering the graph structure} 
We first evaluated the performance of definition generation by Graphex. We compared Graphex with baselines that do not consider the graph structure (\textbf{Table \ref{tab:result}}). We found that Graphex, which uses both the definition graph and the terminology graph, obtained the best performance on all six metrics. The improvement is most prominent against baselines that do not use the graph structure. For example, Graphex obtained 34.35 BLEU1 score, which is 7.65\% and 59.62\% higher than Transformer and Seq2seq. Moreover, we observed decreased performance when only the terminology graph (Our Method w/o DG) or the definition graph (Our Method w/o TG) is considered. Despite less superior performance, these two variants are still consistently better than baselines that do not use graphs, confirming the importance of modeling graph structures in definition generation. 

We showed two examples of how the graph can help Graphine generate better definition (\textbf{Table \ref{tab:case}}). In both examples, the true definition of the nearby node is included in the training set, and can thus be used to capture the global semantic. We found that Graphex selectively copied tokens in the true definition of the parent node, leading to a more accurate generation. For example, in the first case, Graphex successfully generated \textit{estuarine water} and \textit{estuary bed below the water column}. In the second case, Graphex propagated \textit{in a given population in a given time period} from the child node, resulting in the correct generation of \textit{in a given population in a given time}. In contrast, the Transformer baseline is not able to generate such detailed information in both examples due to the ignorance of graph structures. Since the pretrained language model and the graph representation method are two important model selections in Graphex, we next leverage Graphex to compare different pretrained language models and graph representation methods, shedding light on future directions in definition generation.

\begin{table}[]
\small
\centering
\resizebox{\linewidth}{!}{
\begin{tabular}{p{18mm}|p{9mm}p{9mm}p{9mm}p{9mm}p{10.5mm}p{9mm}}
\hline
\textbf{Pretrain}&\textbf{BLEU1}&\textbf{BLEU2}&\textbf{BLEU3}&\textbf{BLEU4}&\textbf{METEOR}&\textbf{NIST}\\
\hline
BERT &30.83	&23.72	&19.95	&17.50	&14.70	&1.02\\
RoBERTa &25.12	&18.36	&14.95	&12.73	&12.17	&0.80\\
SciBERT &33.95	&26.55	&22.53	&19.96	&16.21	&\textbf{1.15}\\
PubMedBERT &31.35	&24.31	&20.57	&18.01	&15.12	&1.05\\
BioBERT &\textbf{34.35}	&\textbf{26.97}	&\textbf{22.99}	&\textbf{20.21}	&\textbf{16.57}		&\textbf{1.15}\\
\hline
\end{tabular}}
 \caption{Comparison on the performance of definition generation using different pretrained language models. SciBERT, PubMedBERT, BioBERT are domain-specific pretrained language models.}
 \label{tab:bert}
\end{table}
\subsection{Comparing Pretrained Language Models}
Domain-specific pretrained language models have achieved impressive performance on tasks such as named entity recognition, information extraction and relation extraction in biomedicine~\cite{scibert,lee2020biobert}. One barrier to more thoroughly comparing these pretrained language models is the lack of domain-specific benchmarks. Graphine could be used as a novel domain-specific benchmark in biomedicine. As a proof-of-concept, we compared five pretrained language models, including three biomedical domain-specific models, by using it to generate the local semantic $\boldsymbol l_i$ in eq. \ref{eq:loss} (\textbf{Table \ref{tab:bert}}). We found that domain-specific pretrained language models have consistently better performance than general pretrained language models, which agrees with previous findings on the value of domain-specific language models in biomedicine \cite{ pubmedbert, lee2020biobert, scibert}. Within the three domain-specific pretrained language models, BioBERT and SciBERT obtained the most prominent performance. This might be due to the corpus these two models were trained on, suggesting the possibility to use Graphine to further compare different biomedical corpus \cite{wang2020cord, lo2019s2orc}. 

\subsection{Comparing graph representation methods}
We next sought to compare graph representation methods using a link prediction task based on our dataset. Graphs in our dataset present a hierarchical structure, which poses a unique challenge for graph representation methods. The results are summarized in \textbf{Table \ref{tab:gcn} }. We found that methods that consider the graph structure have overall superior performance, conforming the importance of the graph structure in definition generation. Among all approaches, GCN obtained the best performance. We didn't observe improved performance by embedding graphs into the hyperbolic space, which is contradictory to prior work showing that hyperbolic embedding can better model hierarchical structures \cite{ nickel2017poincar, hgcn}. We attribute this to the more complicated node features in contrast to previous work. In our dataset, node features are text that has arbitrary length and large vocabulary, introducing new challenges to hyperbolic embedding-based methods.
\begin{table}[]
\small
\centering

\begin{tabular}{l|ll}
\hline
\textbf{Model}&\textbf{AUC}&\textbf{AP}\\
\hline
Euclidean&0.8979&0.9307\\
PoincareBall&0.9069&0.9346\\
\hline
HNN&0.9023&0.9211\\
MLP&0.8892&0.9258\\
\hline
GCN & \textbf{0.9493} & \textbf{0.9659} \\
HGCN &0.8996&0.9365\\
GAT &0.8867&0.9179\\
\hline
\end{tabular}
 \caption{Comparison on the performance of link prediction using different graph representation learning methods.}
 \label{tab:gcn}
\end{table}

\begin{figure*}[!t]
\includegraphics[width=1\textwidth]{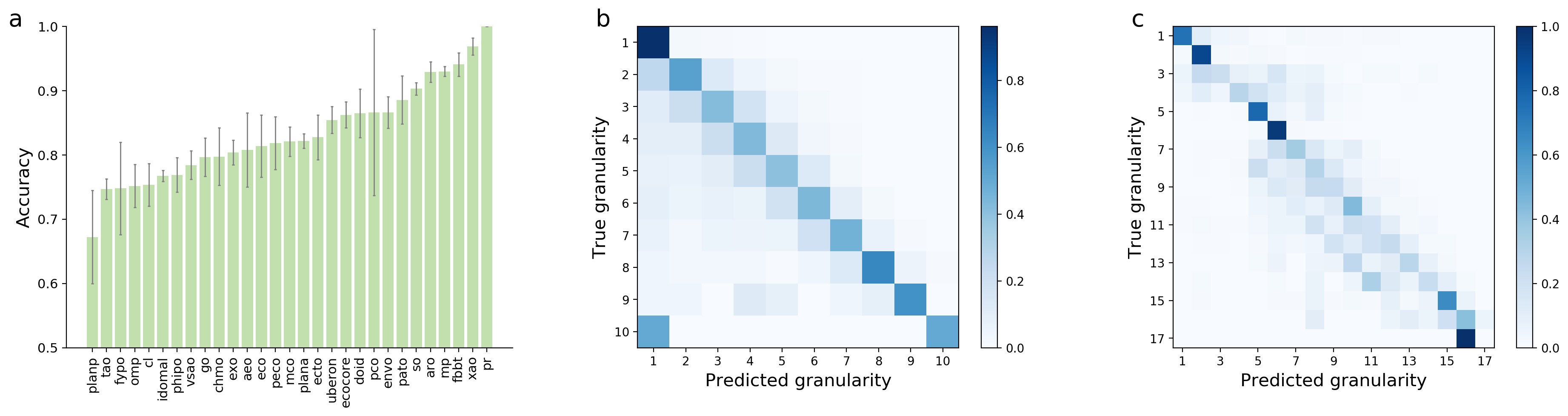}
\caption{Sentence granularity prediction. a, Bar plot showing the accuracy of relative granularity prediction within each DAG. b,c, Heatmaps showing the accuracy of absolute granularity prediction within each DAG (b) and across all DAGs (c).}
\label{fig:7}
\end{figure*}

\subsection{Sentence granularity prediction}
Finally, we exploited Graphine for a novel task of sentence granularity prediction. Measuring sentence semantic similarity is crucial for many NLP tasks. Existing sentence similarity benchmarks only provide binary labels indicating similar or dissimilar \cite{li2006sentence,mueller2016siamese}. In contrast, our dataset is able to characterize the specific granularity of sentences beyond similarity. We define the ground truth granularity of a definition sentence as its depth in the DAG, where a smaller (larger) depth indicates a more coarse-grained (fine-grained) sentence. Based on this granularity benchmark, we define two specific tasks: relative granularity prediction and absolute granularity prediction. Relative granularity prediction aims at predicting which sentence is more fine-grained between two given sentences. Absolute granularity prediction aims at predicting the specific granularity of a given sentence. The incomparable granularity levels from different graphs could introduce systematic bias to comparing sentences from different graphs. To tackle this problem, we first performed a graph alignment among all DAGs using terminologies that appeared in multiple DAGs as anchors. After the alignment, all sentences are associated with a granularity level between 1 and 17, where 1 indicates the most coarse-grained sentence.

To predict the relative granularity, we used the concatenation of the BERT embeddings of two sentences as features to train an multi-layer perceptron (MLP). When comparing sentences within the same DAG, 76\% of graphs obtained an accuracy larger than 0.80 (\textbf{Figure \ref{fig:7}a}). We next examined the accuracy of classifying a pair of sentences from two different DAGs and also observed a good accuracy of 0.81. To predict the absolute granularity, we used the BERT embedding of each sentence as features to train an MLP-based multi-class classifier. We again observed desirable accuracy of 0.71 and 0.81 and Spearman correlation 0.60 and 0.69 within each graph and across all graphs (\textbf{Figure \ref{fig:7}b,c}). In addition to predicting sentence granularity, we envision this new benchmark of sentence granularity could provide deeper insight into evaluating existing sentence similarity models through transforming it from a binary classification task to a ranking task. 
\begin{figure*}[!t]
\includegraphics[width=\textwidth]{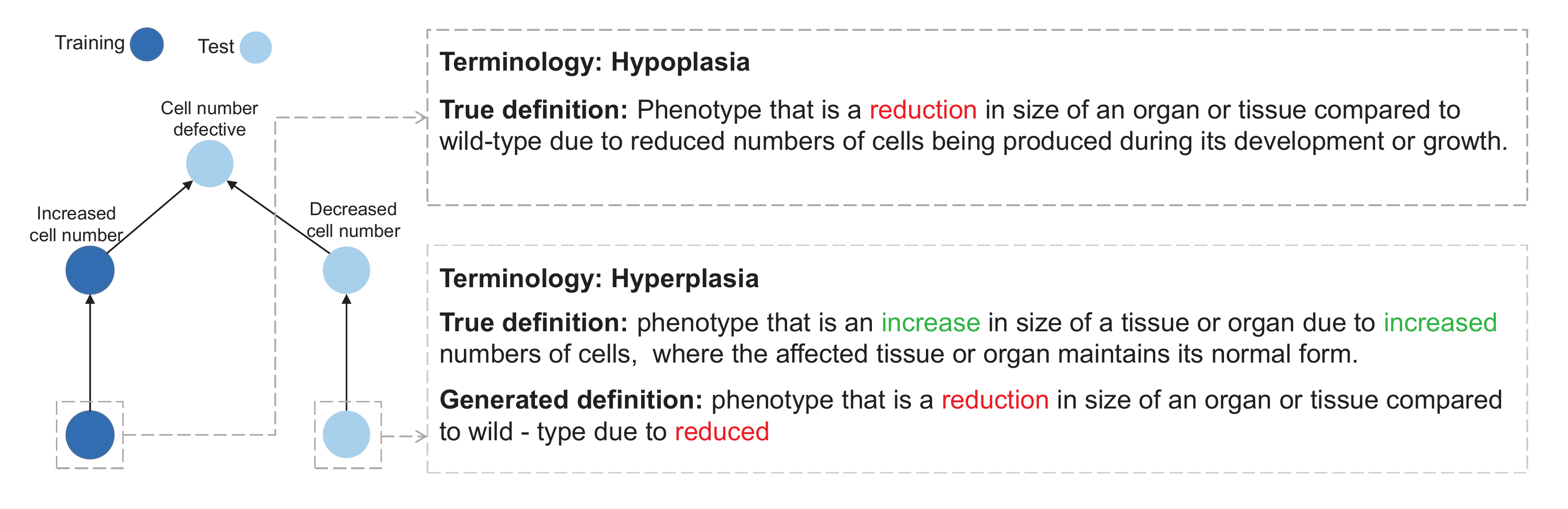}
\caption{A failed generation that cannot be captured by existing evaluation metrics. Graphex generated a sentence that has the opposite meaning to the true definition. }
\label{fig:6.1}
\end{figure*}
\section{Future work motivated by an opposite generation}
Despite the overall improved performance of Graphex, we found that some definitions generated by Graphex present an opposite meaning to the truth definition. We showed one of such example in \textbf{ Figure \ref{fig:6.1} }. Although the definition generated by Graphex for \textit{hyperlasia} matches the true definition well, the generated definition has the opposite semantic meaning (e.g., \textit{reduction}, \textit{reduced}) to the true definition (e.g., \textit{increase}, \textit{increased}). Notably, such failed generations cannot be captured by existing $n$-gram based metrics, leading to artificial improvement. After a closer examination, we found that this opposite generation is caused by using the definition from a cousin node \textit{hypolasia} in the graph. Moreover, existing BERT-based models are not able to effectively associate subword \textit{hypo} (\textit{hyper}) in the terminology with \textit{reduce} (\textit{increase}) in the definition. We plan to explore the possibility of developing faithful generation models \cite{wang2020towards} to address this problem and leave it as an important future work.

\section{Relate Work}
Existing works related to terminology definition mainly focus on definition extraction~\cite{westerhout2009definition,anke2018syntactically,veyseh2020joint,li2016definition} and technology entity recognition~\cite{fahmi2006learning,gao2018adversarial}. Definitions are extracted from different sources, such as Wikipedia~\cite{espinosa2014applying,li2016definition} and scholarly articles~\cite{jin2013mining,spala2019deft}. In contrast to previous work, We study the novel problem of terminology definition generation. Notably, the proposed dataset Graphine can also be used as a new benchmark to evaluate existing approaches on extracting definitions from the free text.

Many scientific literature datasets have been curated for a variety of tasks, such as hypothesis generation~\cite{spangler2014automated}, scientific claim verification \cite{scifact}, paraphrase identification~\cite{vinyals2016show,dong2021parasci,xu2016msr} and citation recommendation ~\cite{saier2019bibliometric}. Paraphrase identification datasets, such as MSCOCO, Quora, MSR, ParaSCI, are most related to our work~\cite{vinyals2016show,dong2021parasci,xu2016msr}. Distinct from these datasets, we focused on a different task (i.e., definition generation) and a different domains (i.e., biomedical domain). 

Graph2text and data2text, which aim at generating text from structured data, have attracted increasing attention~\cite{marcheggiani2018deep,cai2020graph,yao2020heterogeneous,guo2020cyclegt,wang2019deep}. Among them, AMR-to-text Generation and knowledge graph to text generation also consider graph structures. The Abstract Meaning Representation (AMR) represents the semantic information of each sentence using a rooted directed graph, where each edge is a semantic relations and each node is a concept~\cite{song2018graph,zhu2019modeling,mager2020gpt,wang2020better}. Knowledge graph to text generation has advanced tasks such as entity description generation and medical image report by generating text from a subgraph in the knowledge graph ~\cite{cheng2020ent,li2019knowledge}. Despite all considering graph structures, our method generates one sentence for each node on a large directed acyclic graph, whereas AMR-to-text and knowledge graph to text generate sentences for a subgraph or the entire graph.

\section{Conclusion}
We have introduced a novel dataset Graphine for studying definition generation. Graphine includes 2,010,648 terminology definition pairs from three major biomedical databases. Terminologies in Graphine form 227 directed acyclic graphs, which make Graphine a unique resource for exploring graph-aware text generation. We have proposed a graph-aware definition generation method Graphex, which takes the graph structure into consideration. Graphex has obtained substantial improvement against methods that do not consider graph structures. Moreover, we have illustrated how Graphine can be used to evaluate other tasks, including comparing pretrained language models, comparing graph representation learning methods and predicting sentence granularity. Finally, we have analyzed the definition generated by our method and proposed future directions to improve. Collectively, we envision our dataset to be a unique resource for definition generation and could be broadly utilized by other natural language processing applications.

\section*{Acknowledgement}
This paper is partially supported by National Key Research and Development Program of China with Grant No. 2018AAA0101900/2018AAA0101902 as well as the National Natural Science Foundation of China (NSFC Grant No. 62106008 and No. 61772039). 

\bibliography{anthology,custom}
\bibliographystyle{acl_natbib}

\end{document}